\documentclass{article}
 \PassOptionsToPackage{numbers, compress}{natbib}

\usepackage{graphicx}

\usepackage[final]{nips_2018}

\usepackage[utf8]{inputenc} 
\usepackage[T1]{fontenc}    
\usepackage{hyperref}       
\usepackage{url}            
\usepackage{booktabs}       
\usepackage{amsfonts}       
\usepackage{nicefrac}       
\usepackage{microtype}      
\setcitestyle{square}

\title{
Two-stream Convolutional Networks for End-to-end Learning of Self-driving Cars
}

\author{
  Nelson Fernandez\textsuperscript{a,\dag}\\
  \textsuperscript{a} Renault Automotive\\
  \texttt{nelson.fernandez-pinto@renault.com} \\
  \textsuperscript{\dag} Previously at Axionable Labs, Paris France\\
}

\newcommand\blfootnote[1]{%
  \begingroup
  \renewcommand\thefootnote{}\footnote{#1}%
  \addtocounter{footnote}{-1}%
  \endgroup
}

\begin{document}
\blfootnote{\dag This work has been supported by  Axionable Labs, Paris 75003, France.}

\maketitle
\begin{abstract}
We propose a methodology to extend the concept of Two-Stream Convolutional Networks to perform end-to-end learning for self-driving cars with temporal cues. The system has the ability to learn spatiotemporal features by simultaneously mapping raw images and pre-calculated optical flows directly to steering commands. Although optical flows encode temporal-rich information, we found that 2D-CNNs are prone to capturing features only as spatial representations. We show how the use of Multitask Learning favors the learning of temporal features via inductive transfer from a shared spatiotemporal representation. Preliminary results demonstrate a competitive improvement of 30\% in prediction accuracy and stability compared to widely used regression methods trained on the Comma.ai dataset.
\end{abstract}

\section{Introduction}
Decision making in the spatiotemporal domain is a key issue for autonomous driving systems \cite{DecisionMaking}. The current paradigm is the implementation of a Convolutional Neural Network (CNN) to perform direct steering commands regression from raw images \cite{Pilotnet}. A forward-facing camera records video footage of the road, which is tagged with the vehicle's cinematic measurements, such as speed, wheel angle and acceleration. The data is then used to train the CNN regressor with a supervised learning approach. This system has proven to be efficient on extracting spatial cues \cite{Pilotnet2} and achieves respectable accuracy. The main limitation is that frames are analyzed individually without taking into account previous actions. Therefore, any temporal information is lost due to this abstraction.

Following its introduction by Symonyan and Zisserman (2014) \cite{Simonyan}, Two-Stream Convolutional Networks have become widely adopted as the preferred method for action recognition in videos. The main advantage is the exploitation of the temporal dynamics captured by the optical flow between adjacent frames. In this case, the temporal stream is prone to decoding features instead of making an estimation of the motion field. To the best of our knowledge, the use of this architecture in regression tasks, especially autonomous steering, has not been studied in depth. 

It is widely believed that the use of temporal features together with spatial characteristics could lead to better model predictions. Currently, the main research efforts are focused on the use of Recurrent Neural Networks (RNN) \cite{Eraqi} and 3D-Convolutional Neural Networks (3D-CNN) \cite{3D}. In contrast to intuition, the use of 2D-CNNs could bring some advantages compared to RNNs, in particular, a faster and more straightforward training. Also, 2D-CNNs are less likely to overfit when performing temporal tasks \cite{overfit}. On the other hand, 3D-CNNs must perform an implicit estimation of the optical flow while trying to learn temporal features through the depth of the input image tensor, which makes training difficult.

 The principal motivation of this work is to demonstrate that Two-Stream Convolutional Networks are a robust alternative to perform end-to-end learning with temporal cues. This article describes the preliminary results tested on Comma.ai's dataset \cite{Comma} compared to other regression methods.

\section{Methods}
\label{gen_inst}
Based on the Two-Stream Convolutional Network architecture, it is possible to design a custom self-driving system that takes advantage of spatial and temporal features using 2D-CNNs. This system is composed of two visual streams taken from a single front-facing camera and its optical flow calculated from the previous image. The optical flow can be estimated efficiently in almost real time using the current accelerated computing technology. In addition, specialized hardware can be integrated to obtain live optical flow streams. A recording device uses the controller area network (CAN) bus to collect steering wheel commands directly from the vehicle's computer. This data is used to train a regressor, which provides a prediction of the required steering angle based on inputs. The network output is decoded and sent to a proportional–integral–derivative controller (PID), to finally steer the position of the wheels. Figure 1 shows an overview of the proposed self-driving car system. 

\begin{figure}[h!]
  \centering
  \includegraphics[width=0.85\textwidth]{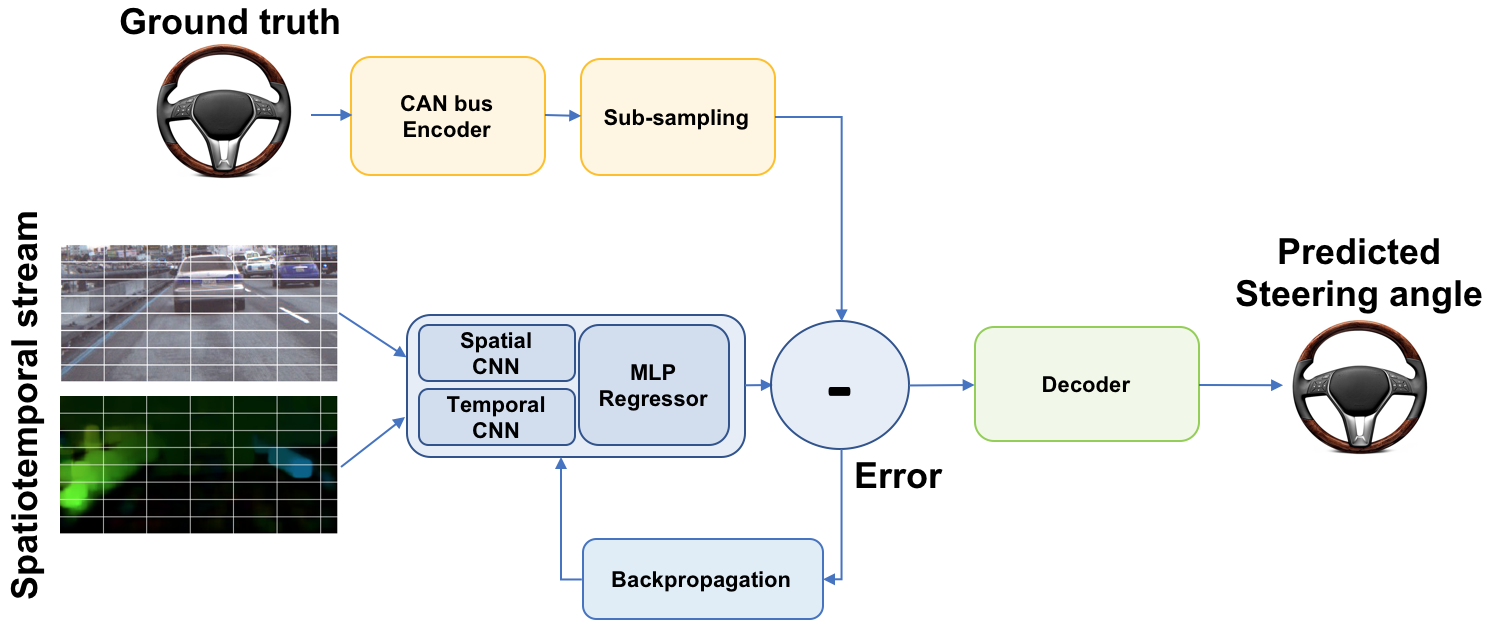}
  \caption{Proposed Two-Stream self driving navigation system.}
\end{figure}

Figure 2 shows the proposed network architecture that incorporates key elements of Two-Stream Convolutional Networks and 2D-CNNs to steering angle regression. The ensemble is composed by two identical CNN branches similar to \cite{Inception}. Global spatial pooling is applied in the last convolutional layers, aggregating information from all dimensions into single real values. This global aggregation considerably reduces the number of parameters and decreases the risk of overfitting \cite{GAP}. The resulting embeddings are merged with a fusion layer of type element-wise multiplication, forming a shared spatiotemporal representation. A multi-layered perceptron (MLP) with linear output neurons performs the regression. 
Following the Two-Stream Networks logic, the first branch is intended to learn spatial features extracted from raw images \cite{Pilotnet2}. The second learns temporal characteristics from the estimated relative displacement of objects between frames. 

Based on the inputs, it is possible to train the network to simultaneously predict the current and previous state of the target variable using Multitask Learning (MTL). As the goal of architecture is to capture short-term temporal dependencies, we assume that on the horizon studied, the inputs provide the information needed to perform the regression. The main motivation of performing MTL is to favor the learning of temporal features via inductive transfer \cite{Caruana}. By consequence, the auxiliary target is dropped during the inference step. The architecture has about 45 million parameters and is trained using backpropagation.

\begin{figure}[h!]
  \centering
  \includegraphics[width=1\textwidth]{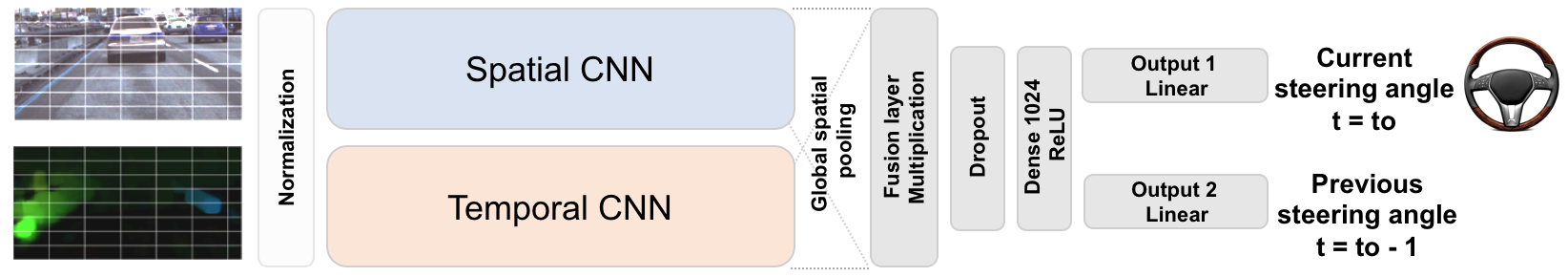}
  \caption{Proposed Two-Stream Convolutional Network architecture.}
\end{figure}

The model was trained with 6.2 driving hours from the Comma.ai dataset \cite{Comma}, corresponding to approximately 220K image/optical flow pairs of size 160x320x3 sampled at 10 Hz. The optical flow was calculated using the Farnebäck's \cite{Optical} implementation available in OpenCV. The estimated motion field was mapped to a standardized RGB color wheel. Dropout and data augmentation strategies, including spatial and photometric transformations were applied to avoid overfitting. Due to the small size of the training set, this work is more aimed at obtaining a benchmark rather than training an autopilot ready for production.

A two-step training strategy was used, with 60\% of the training data for feature learning, and 40\% for fine-tuning. During the first step, the CNN branches are trained independently, the weights obtained are used as initialization of the second stage. In the second step, we fine-tune the architecture by focusing on the MLP. We set the following hyper parameters: Loss function of type Mean Squared Error (MSE), Adam optimizer \cite{Adam} with learning rate 1e-4 for feature learning and 0.5e-4 for fine tuning, and a batch size of 64 for feature learning, and 8 for fine tuning. The model trained for 30 epochs in the feature learning step and for 1 epoch for fine tuning. The architecture was implemented using the Keras functional API running on a machine with a single NVIDIA M6 GPU and 120Gb of RAM. 

The test set comprised 56 minutes of video taken from the second and third files of the Comma.ai dataset, approximately 33K images sampled at 10Hz. This was done to prevent data leakage due to similar frames in the training and test partitions in accordance with \cite{Eraqi}. As no examples of these folders were included in the training set, the regression task becomes harder, measuring the generalization capabilities of the model.

The main evaluation metric was the Root Mean Squared Error (RMSE), that measured the accuracy of the steering angle regression. We also used the whiteness \cite{Eraqi}, that measured the smoothness of the predictions over time. This quantity is based on the square of the first time derivative calculated according to expression (1). The whiteness reflects the temporal dynamics of the steering task. Low values mean smooth steering, and therefore a more comfortable driving experience for users. 

\begin{equation}\label{(whiteness)}
Whiteness =\sqrt{\frac{1}{D}\sum_{i=1}^{D}\frac{\partial P_{t_{i}}}{\partial t}^{2}}[\frac{degrees}{time}]
\end{equation}

\section{Results}
\label{headings}
The first experiment is intended to evaluate the effect of using optical flows instead of raw images to train single 2D-CNNs. For this, we trained two widely used architectures to perform steering angle regression directly from images and optical flows respectively. The main idea is to assess the capability of CNNs to learn temporal features from the displacement of objects between consecutive frames. The results of this experience are shown in table 1.

\begin{table}[hbt!]
  \caption{Test set RMSE and whiteness of single-CNN architectures.}
  \label{simple-table}
  \centering
  \begin{tabular}{ccccc}
    \toprule
    \begin{tabular}{lllll} & {Trained with raw images}  &  & {Trained with optical flows} &\\
    \cmidrule(r){2-5}
    Model & RMSE[degrees]& Whiteness & RMSE [degrees]& Whiteness\\
    \midrule
    Simple-CNN  \cite{Pilotnet} & 20.55  & 8.84 & \textbf{14.21}  & 8.74 \\
    Inception V3 & 17.76  & 7.05 & \textbf{12.58}  & 7.15 \\
    \bottomrule
    \end{tabular}
  \end{tabular}
\end{table}

The above results show that whichever CNN architecture is utilized, the optical flow produces significant accuracy gains in the test set compared to raw images. We also see some improvement associated to the architecture sophistication of Inception V3 \cite{Inception} compared to simple CNNs. These results suggest that the object displacement representation is convenient to learn specific features improving generalization. The fact that the whiteness did not improve, implies that the networks are learning features only as spatial representations.

In the second experiment, we evaluate the proposed architecture alongside other widely-used steering angle regression methods. Table 2 shows that Two-Stream Convolutional Network outperformed state-of-the-art 2D-CNN models in prediction accuracy and stability, with a whiteness closer to the human driver. 

\begin{table}[ht!]
  \caption{Test set steering wheel angle regression error.}
  \label{sample-table}
  \centering
  \begin{tabular}{lll}
    \toprule
    \cmidrule(r){1-2}
    Architecture     & RMSE [degrees] & Whiteness [degrees/time]\\
    \midrule
    Human driver & N/A  & 4.36     \\
    Comma.ai 2D-CNN & 23.99  & 9.81     \\
    Nvidia PilotNet & 20.55 & 9.23      \\
    Inception V3    & 17.76 & 7.05      \\
    \textbf{Two-Stream Network} & \textbf{12.52}  & \textbf{4.97}      \\
    \bottomrule
  \end{tabular}
\end{table}

Figure 3 shows the scatter plot of the test set instances in paired coordinates of instantaneous whiteness (x-axis) and steering angle (y-axis), along with predictions of three different CNN regression models. As the whiteness is proportional to the square of the first time derivative, the figure is a spatiotemporal representation of the steering wheel angle task. The proposed architecture (right) helps to resolve the bias in the spatial and temporal axis of individual CNN streams trained with raw images (left) and optical flows (center). This improvement is achieved by the incorporation of a shared spatiotemporal representation and MTL from a time-related task. The learning of the previous steering angle as an auxiliary task, produces bias correction in the temporal domain, reducing the whiteness. This filtering effect would be difficult to achieve using only single-target individual CNNs.

\begin{figure}[h!]
  \centering
  \includegraphics[width=1\textwidth]{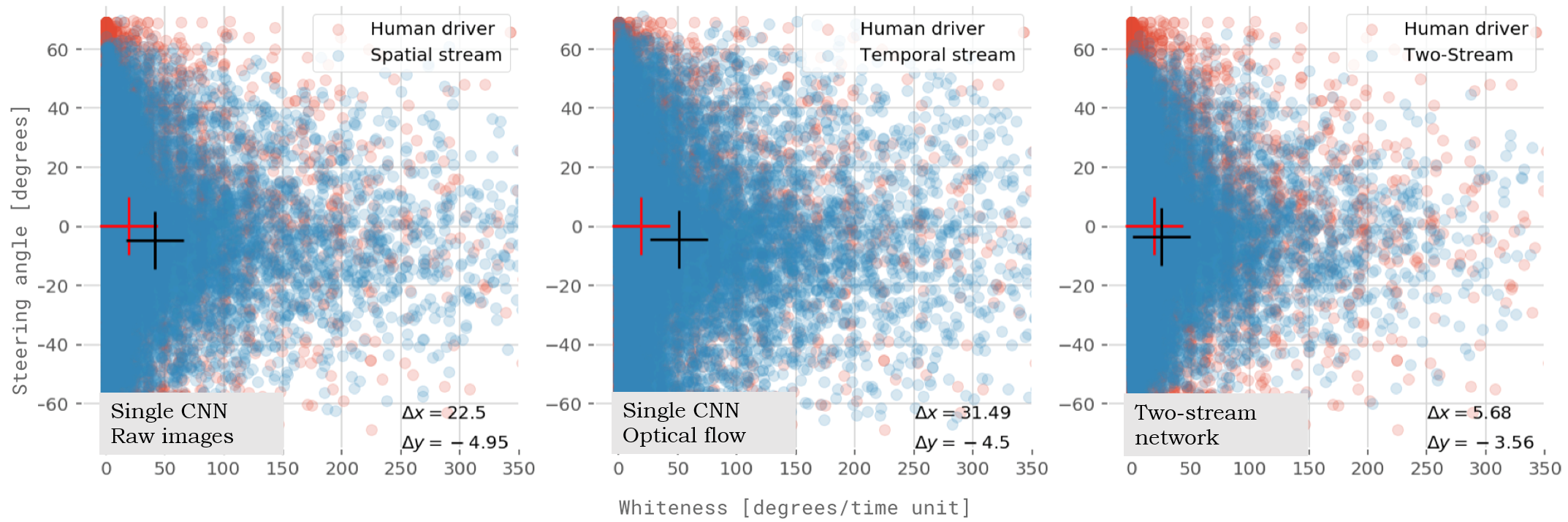}
  \caption{Test set model bias of a spatial stream (left), temporal stream (center) and the proposed architecture (right).}
\end{figure}

\section{Discussion}
\label{others}
The main contribution of this work is to demonstrate that Two-Stream Convolutional Networks are a robust alternative to augment end-to-end learning with spatiotemporal cues. We found that the use of optical flows instead of raw images can significantly increase the accuracy of current self-driving systems. The main reason for this is that optical flows provide a convenient representation that can be easily learned by CNNs. However, the use of optical flows alone is not sufficient to incorporate temporal dependencies, as CNNs are biased to learning characteristics as spatial representations.

The proposed Two-Stream architecture learns how to effectively combine spatial and temporal information from a shared spatiotemporal representation that encodes the relative movement of objects between two consecutive frames. The introduction of a related auxiliary task (the previous steering angle) guides the network to discover short-term temporal dependencies via inductive transfer. Then, an important bias correction is achieved in the spatial and temporal domains improving generalization. The results show a competitive performance increase in prediction accuracy and stability of 30\%, when compared to widely-used regression methods trained on the Comma.ai data set. The next steps of this work will focus on the study of MTL for autonomous driving regression tasks.


\end{document}